\documentclass[11pt,a4paper]{article}
\usepackage{times,latexsym}
\usepackage{url}
\usepackage[T1]{fontenc}
\usepackage{paralist}
\usepackage{graphicx}
\usepackage[utf8]{inputenc}
\usepackage{inconsolata}

\usepackage{zshi}
\usepackage{caption}
\usepackage{subcaption}
\usepackage{algorithm}
\usepackage[noend]{algorithmic}
\usepackage{microtype}
\usepackage{arydshln}
\usepackage{makecell}
\usepackage{tabularx}
\usepackage{hhline}
\newcommand\setrow[1]{\gdef\rowmac{#1}#1\ignorespaces}
\usepackage{graphicx}

\usepackage[acceptedWithA]{tacl2021v1}

\usepackage{xspace,mfirstuc,tabulary}

\title{Automatic Engineering of Long Prompts}

\author{
Cho-Jui Hsieh, Si Si, Felix X. Yu, Inderjit S. Dhillon\\
Google Inc.\\
\texttt{\{cjhsieh, sisidaisy, felixyu, isd\}@google.com}
}
\date{}
\begin{document}
\maketitle

\begin{abstract}
Large language models (LLMs) have demonstrated remarkable capabilities in solving complex open-domain tasks, guided by comprehensive instructions and demonstrations
provided in the form of prompts. However, these prompts can be lengthy, often comprising hundreds of lines and thousands of tokens, and their design often requires considerable human effort. Recent research has
explored automatic prompt engineering for short prompts, typically consisting of one or
a few sentences. However, the automatic design of long prompts remains a challenging
problem due to its immense search space. In
this paper, we investigate the performance
of greedy algorithms and genetic algorithms
for automatic long prompt engineering. We
demonstrate that a simple greedy approach
with beam search outperforms other methods
in terms of search efficiency. Moreover, we
introduce two novel techniques that utilize
search history to enhance the effectiveness of
LLM-based mutation in our search algorithm.
Our results show that the proposed automatic
long prompt engineering algorithm achieves
an average of 9.2\% accuracy gain on eight
tasks in Big Bench Hard, highlighting the
significance of automating prompt designs
to fully harness the capabilities of LLMs.
\end{abstract}

\section{Introduction}

\begin{figure*}[th]
    \centering
    \includegraphics[width=01.0\textwidth]{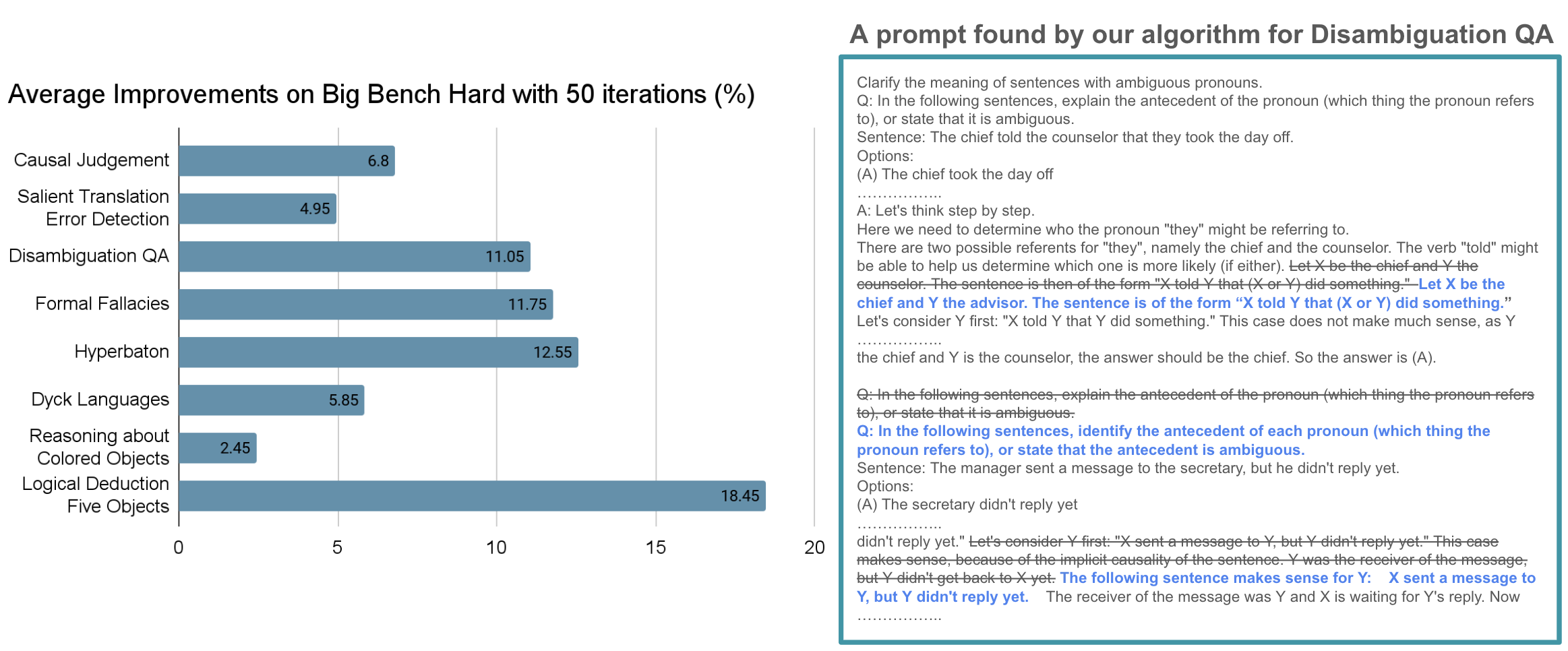}
    \caption{{\bf Left panel}: the average accuracy improvements of the proposed method over 3 runs with 50 iterations on the training set. We report the accuracy on the whole set (including training and test sets). More detailed results can be found in Table~\ref{tab:main_results}. {\bf Right panel}:
    An example of a long prompt in BBH (Disambiguation task) which consists of an instruction, several demo examples and chain-of-thought reasoning. We show that by rewriting a few selected sentences in this long prompt with our proposed method, we can improve the accuracy by more than 10\%. }
    \label{fig:example_1}
\end{figure*}

Large language models (LLMs) have exhibited remarkable capabilities when trained on large datasets, demonstrating the ability to comprehend complex and lengthy instructions for diverse tasks without the need for fine-tuning~\cite{wei2022emergent,brown2020language,chowdhery2022palm,ouyang2022training}. Several prompt design principles have emerged in recent years, suggesting that incorporating more complex instructions, demonstrations, and chain-of-thought reasoning into prompts can boost the performance on challenging tasks~\cite{brown2020language,wei2022chain}, including those involving mathematical problem-solving~\cite{cobbe2021training} and reasoning~\cite{suzgun2022challenging,srivastava2022beyond}. However, effective prompts for tackling complex tasks often contain thousands of tokens, posing challenges in designing and optimizing them. Figure~\ref{fig:example_1} demonstrates a long prompt for a task in Big Bench Hard~\cite{suzgun2022challenging}, which contains an instruction and several demos, each with human-written chain-of-thoughts. 

Numerous studies have demonstrated the sensitivity of LLMs to prompts, revealing that minor modifications, such as adding or removing a few tokens or rephrasing the prompt, can significantly impact LLM performance~\cite{liu2023pre,zhu2023promptbench,jiang2020can}. Therefore, prompt design has become a labor-intensive endeavor. Further complicating the matter is the rapid evolution of LLMs, rendering prompts crafted for a specific LLM ineffective when applied to a newer version of LLM. This highlights the need for automatic prompt engineering techniques.

While automatic prompt engineering has been studied recently, existing research~\cite{deng2022rlprompt,xu2022gps,guo2023connecting,fernando2023promptbreeder} focuses on optimizing short instructions with one or a few sentences. These methods either evolve prompts by searching for word replacements or utilize LLMs to rewrite the entire prompt~\cite{xu2022gps,fernando2023promptbreeder,guo2023connecting,yang2023large}. It is challenging to apply them for evolving a long prompt like Figure~\ref{fig:example_1}. For long prompts, word replacement-based search faces an immense search space, while rewriting the entire prompt using a single LLM query is extremely difficult. Although it is possible to apply these methods to individual sentences within a prompt, we have observed that tuning a single sentence may not sufficiently improve the performance of long prompts  (see Section~\ref{sec:exp}). Therefore, {\it how to automatically find better long pompts and what's the potential performance gain by tuning long prompts} remain unanswered in the literature. 


The main operation in our automatic prompt engineering technique is to replace a sentence by a semantically equivalent sentence, which can be done by queries to LLMs. By comparing the pros and cons of greedy algorithm and genetic algorithm, two representative algorithms for discrete optimization, we show that a simple greedy algorithm with beam search is more effective. Further, for long prompt engineering it is important to identify which sentence to change, and to what direction. We thus propose two novel techniques that utilize search history to enhance the effectiveness of LLM-based mutation in our algorithm.

Our contributions can be summarized below: 
\begin{itemize}
    \item To the best of our knowledge, this paper presents the first formal discussion of automatic long prompt engineering, demonstrating substantial performance gains on multiple tasks. 
    \item We propose a greedy algorithm with beam search that can rapidly optimize prompts. Furthermore, a novel guided mutation method is introduced to enhance convergence. 
    \item We conducted experiments on the Big Bench Hard (BBH) benchmark~\cite{suzgun2022challenging,srivastava2022beyond}, where prompts comprise thousands of tokens, including instructions and chain-of-thought reasoning. Our results demonstrate that the proposed automatic long prompt engineering method significantly enhances performance. Notably, we achieved an average of 9.2\% absolute accuracy improvements on 8 tasks selected from BBH, as shown in Figure~\ref{fig:example_1}, with only 50 evaluations on the training set.  
\end{itemize}

\section{Related Work}

The remarkable ability of large language models (LLMs) to perform complex tasks without fine-tuning through prompting has significantly broadened their applicability. As a result, designing effective prompts to fully exploit the capabilities of LLMs has become an important topic. Many principal ways for prompt design have been studied recently~\cite{reynolds2021prompt,brown2020language,wei2022chain,wang2022self,wang2023plan}.  For instance, a well-designed prompt may include a system prompt (e.g., "you are an AI programming assistant"), an instruction prompt outlining the task, multiple contextual examples, and a chain-of-thoughts reasoning section that explains the step-by-step thought process behind the examples. By incorporating these elements, effective prompts often have tens of sentences and thousands of tokens.

An orthogonal line of previous work has explored soft-prompt tuning, a technique that optimizes prompts within a continuous embedding space using standard continuous optimization algorithms~\cite{lester2021power,zhang2021differentiable,wang2022preserving}. While capable of achieving satisfactory performance, soft-prompts lack interpretability and cannot be applied via LLM APIs. Additionally, these parameter-efficient fine-tuning methods demand large training sets, making them unsuitable for applications with limited data, such as those with only tens or hundreds of samples.

Given the limited availability of training data (<1000 samples), our focus lies in exploring strategies for optimizing hard prompts, which are semantically equivalent to the original prompts but yield superior performance. In the context of automated prompt engineering, the literature considers two primary settings. The first setting, which aligns with our work, assumes the existence of an initial human-crafted prompt and aims to refine or improve it to achieve enhanced performance. Several discrete search algorithms have been proposed for this setting: \cite{xu2022gps} employs a genetic algorithm for prompt tuning, utilizing back translation, cloze tasks, and sentence continuation to mutate the initial instruction. More recently, \cite{fernando2023promptbreeder,guo2023connecting} proposed leveraging LLMs for mutation and crossover operations in evolutionary searches, while \cite{yang2023large} demonstrated the optimization capabilities of LLMs in generating improved prompt variations based on previous fitness scores. However, these prompt evolution techniques are designed for short sentences or paragraphs within a long prompt. For example, many of them try to evolve only the instruction part or  the sentence ``Let's think step by step'' in the prompt. Our work aims to provide complete freedom to evolve the entire long prompt, opening up more avenues for improvement but also introducing challenges in determining how and where to change the original prompt.

Another setting focuses on automatic prompt generation without a pre-existing prompt. \cite{honovich2022instruction} demonstrated the ability of LLMs to generate brief task descriptions when provided with input-output pairs. Building upon this technique and incorporating random search, \cite{zhou2022large} proposed an automatic prompt engineering (APE) algorithm capable of generating prompts from given pairs. \cite{pryzant2023automatic} explored the utilization of input-output pair feedback to refine instructions, while \cite{chen2023instructzero} developed a continuous relaxation approach employing Bayesian optimization for search within a continuous space.

\section{Proposed Method}

In this paper, we address the challenge of automatic long prompt engineering for language models. Given a language model $\mathcal{L}$ and an initial prompt $p_\text{init}$ designed by human for a particular task, our goal is to
refine the prompt to achieve superior performance. To enhance the prompt, we are provided with a limited training set (e.g., 100 input-output pairs) $\{(x_i, y_i)\}_{i=1}^n$ for performance evaluation. Specifically, for each sample we conduct prediction by $\mathcal{L}([p, x_i])$ and assess its agreement with the corresponding ground truth label $y_i$. We define $\text{score}(p)$ as the performance metric of prompt $p$ on the training set. We consider the scenario where the number of available samples $n$ is limited, which is a common situation where individuals lack sufficient data for model fine-tuning but can still utilize the data to design an improved hard prompt. We will evaluate generalization performance on a hold-out test set that is not used in prompt search. 

In this section, we will first 
introduce the search space of our algorithm and an exploration of how to use LLM to navigate this space (Section~\ref{sec:space}).
We then delve into the proposed greedy algorithm with beam search, highlighting its advantages over both vanilla greedy algorithms and genetic algorithms in Section~\ref{sec:framework}.
Finally, Section~\ref{sec:guided} 
shows how to leverage search history to guide the proposed framework to improve its performance.

\subsection{Search space}
\label{sec:space}
Our goal is to generate a new prompt that is semantically similar 
to the original prompt while achieving enhanced performance. 
Further, we avoid introducing non-interpretable tokens, such as adversarial triggers~\cite{zou2023universal}. These restrictions offer two advantages: 1) the prompts discovered by our algorithm are interpretable, facilitating the verification that the prompt still performs the intended task, and 2) since we are only provided with a limited set (e.g., 100) of training samples, constraining the search space can mitigate the issue of overfitting.

\begin{figure}
    \centering
    \includegraphics[width=.4\textwidth]{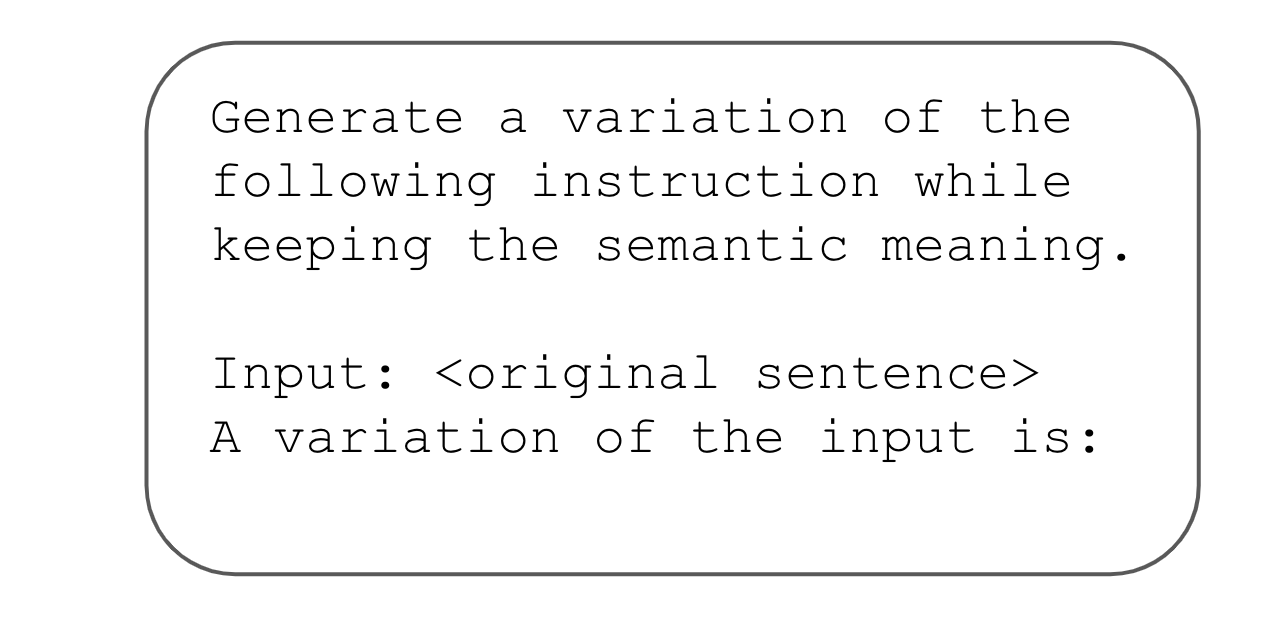}
    \caption{The vanilla LLM-Mutator used in our search. }
    \label{fig:LLM-mutator}
\end{figure}

To conduct prompt search within this  space, we decompose the long prompt into $m$ individual sentences:
\begin{equation*}
p = [s_{(1)}, s_{(2)}, \dots, s_{(m)}].
\end{equation*} 
We then allow each sentence to be rephrased while preserving its semantic meaning. Since LLMs excel at sentence rephrasing, particularly when provided with prompts like "Generate a variation of the following instruction while keeping the semantic meaning," we employ an {\bf LLM-Mutator} equipped with such a prompt to identify alternative formulations for each sentence. The prompt we used for vanilla LLM mutator can be found in Figure~\ref{fig:LLM-mutator}. An improved version of LLM-Mutator will be introduced later in Section~\ref{sec:guided}. 

\subsection{Search algorithm}
\label{sec:framework}

A straightforward approach to conducting the search using the LLM-Mutator is the following greedy algorithm. At each iteration, we randomly select a sentence, denoted by $s_{(i)}$, and utilize the LLM-Mutator to generate an alternative sentence $s_{(i)}'$. We then replace $s_{(i)}$ by $s_{(i)}'$ in the old prompt $p$ if the new resulting prompt improves the performance. However, we observed that this vanilla greedy approach can be easily  trapped in local optima. In our problem, the training set is so limited that sometimes a detrimental modification is not reflected in the training score. As a result, unfavorable edit are sometimes accepted due to insufficient evaluation, which hurts the prompt's future improvement.

\begin{figure*}[ht]
\centering
\begin{tabular}{cc}
\includegraphics[width=.4\textwidth]{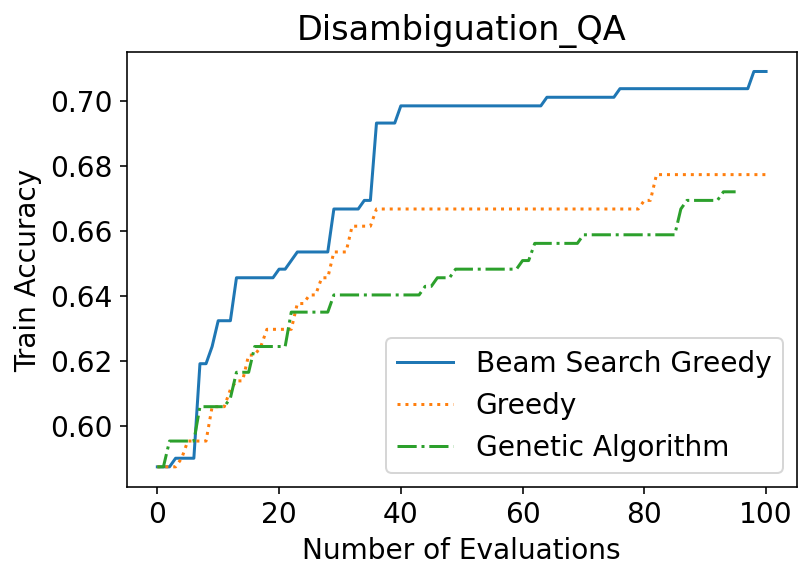} &
\includegraphics[width=.4\textwidth]{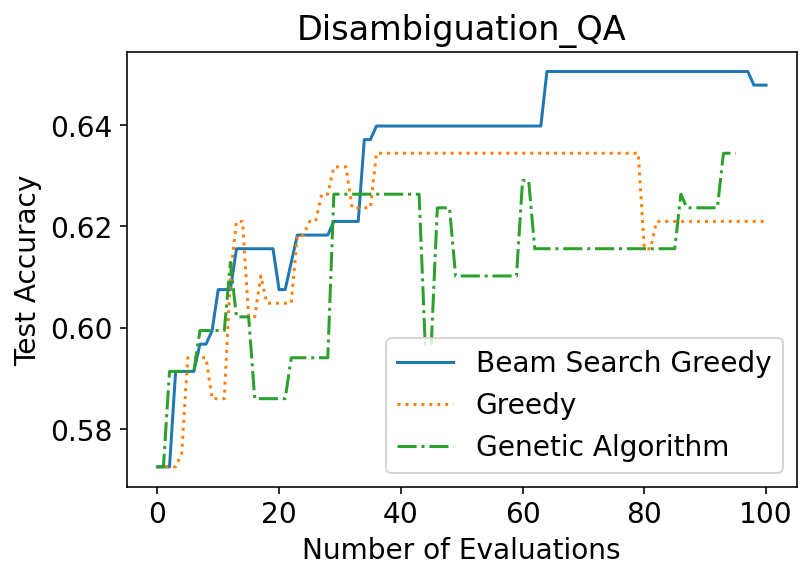} \\
\includegraphics[width=.4\textwidth]{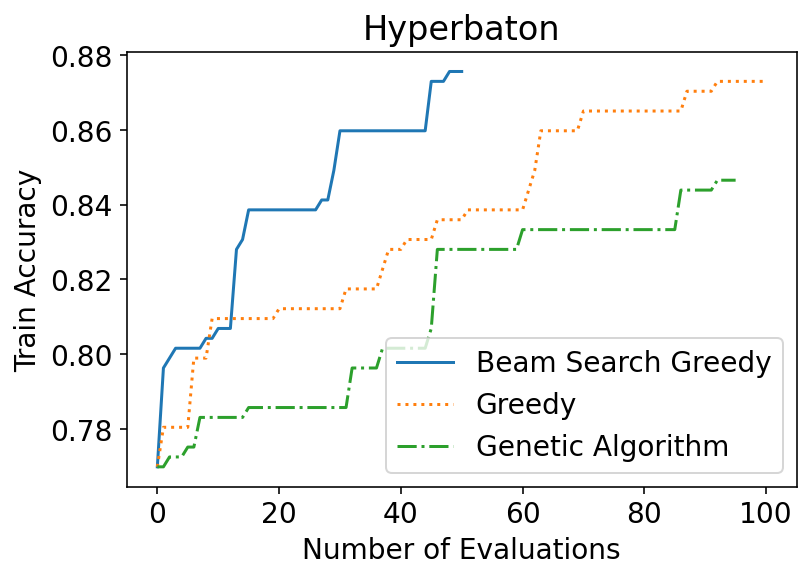} &
\includegraphics[width=.4\textwidth]{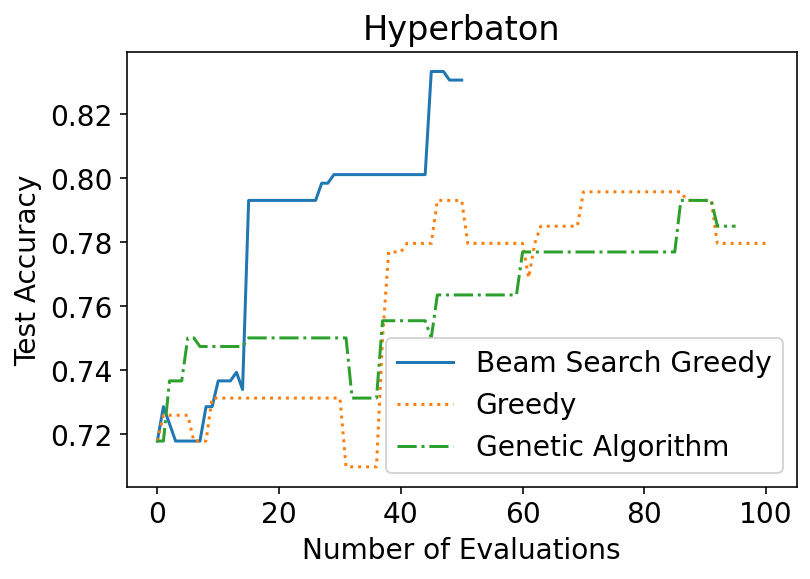}
\end{tabular}
\caption{Comparison between beam-search greedy, greedy, and genetic algorithm. This is the average results over three runs. Beam-search greedy outperforms both greedy algorithm and genetic algorithm.
}
\label{fig:greedy_compare}
\end{figure*}

To address this issue, we propose conducting a beam search by maintaining a pool of $k$ top-performing prompts, denoted as  $p^*_1, p^*_2, \dots, p^*_k$. At each iteration, we randomly select one of these $k$ prompts and  refine the chosen prompt. We then evaluate this new candidate and maintain the top-$k$ prompt pool. This approach ensures that even with the introduction of detrimental edits, recovery from such errors is still possible as we retain not only the top prompt. Our experiments reveal that this approach leads to significantly improved training and test performance compared to the pure greedy algorithm, as demonstrated in Figure~\ref{fig:greedy_compare}. 


It is worth noting that our method is closely related to the Genetic Algorithm (GA). GA is a widely recognized method for discrete optimization involving black-box functions. Assuming $P = \{p_1, \dots, p_k\}$ represents the solutions in the current pool, GA applies mutation and crossover operations to this pool to generate a set of newly proposed candidates  $P' = \{p'_1, \dots, p'_k \}$. The fitness scores (performance of the solutions) of these newly proposed candidates are then evaluated. Subsequently, only the top $k$ solutions in $P\cup P'$ are retained before proceeding to the next iteration.


While GA possesses a high exploration capability, 
its convergence rate is considerably slower during the initial stages. Given the vastness of our search space and the reasonably good quality of the human-written initial prompt, a majority of mutations and crossovers performed in the first few iterations yield poorly performing candidates. Evaluating each of these candidates is computationally expensive, even though it can enhance solution diversity.

Our algorithm is effectively a ``greedy'' version of GA. We maintain a candidate set of size $k$ and update the candidate pool immediately upon generating each new candidate, rather than waiting for the evaluation of the entire generation of offspring before updating the pool. As a result, our solution pool remains more up-to-date, leading to faster convergence compared to GA. Figure~\ref{fig:greedy_compare} further verifies this observation, demonstrating the slower initial growth of GA's learning curve.

\subsection{History-guided search}
\label{sec:guided}
Randomly rephrasing a sentence in a prompt is akin to conducting a random mutation within the space of semantically equivalent sentences, potentially requires a significant amount of trials to obtain a good solution. In this subsection, we demonstrate how the search history can be employed to guide the mutation in a more purposeful direction. 

\paragraph{Guided mutation for a single sentence.}
At iteration $T$, we can denote the search history as $\{(s_t^\text{before}, s_t^\text{after}, r_t)\}_{t=1}^{T-1}$, where $s_t^\text{before}$ is the sentence before mutation, $s_t^\text{after}$ is the sentence after mutation, and the reward $r_t$ indicates  the change in the score ($r_t$ is positive if the mutation enhances the performance,  and negative if it deteriorates performance). Assume $s_T$ is the current sentence that is selected for mutation, we can then use the history to guide the mutation towards more positive reward. For example, if we know rephrasing the sentence ``So, it is true that Lesley is a great-grandfather of Leroy'' to ``Therefore Leslie is Leroy's great-grandfather.'' can improve the performance, then for another sentence ``So, it is true that everyone who is an ancestor of Dana is a stepbrother of Brian'' we may want to rewrite it as ``Therefore everyone that is an ancestor of Dana is Brian's stepbrother.'' 

\begin{figure}
    \centering
    \includegraphics[width=.48\textwidth]{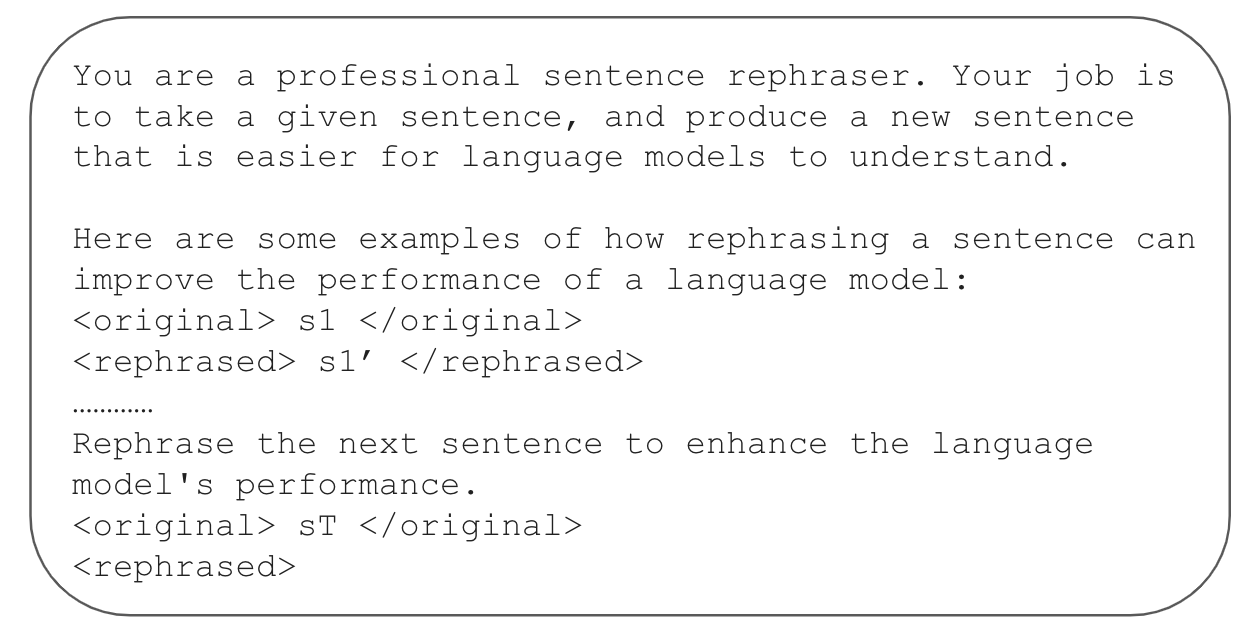}
    \caption{The guided LLM-mutator used in our search. }
    \label{fig:LLM-rephraser}
\end{figure}

It has been shown that LLMs are able to learn from in-context examples~\cite{zhang2021differentiable}. Therefore, we propose to let the LLM-Mutator in-context learn from the history. This can be done by listing history in the prompt: if $r_t>0$, we include an in-context example $s_t^\text{before} \Rightarrow s_t^\text{after}$, and if $r_t<0$ we include $s_t^\text{after} \Rightarrow s_t^\text{before}$. The prompt for LLM-Evolver is shown in Figure~\ref{fig:LLM-rephraser}.

Since the history can be long and it has been shown that putting too many in-context learning examples can be harmful for the performance, we  only retrieve a small set of relevant history entries when rephrasing a sentence. Specifically, when evolving $s_T$, we compute the similarity between $s_T$ and each $s_t^\text{before}$, which is calculated by $\phi(s_T)^T \phi(s_t^\text{before})$ and $\phi$ is a sentence encoder (we use a T5~\cite{raffel2020exploring} sentence encoder in our experiments). We then only select entries that pass a certain threshold to include as in-context examples. 

\paragraph{Guided sampling for sentence selection.}
A challenge in automatic long prompt engineering lies in identifying which sentences  in the prompt should be modified to improve performance. Our experiments demonstrate that altering only a few sentences in a long prompt can significantly enhance its effectiveness, but the changes need to be at right places. It is thus  beneficial to bias the sampling distribution towards selecting more impactful sentences for modification.

We model sentence selection as a contextual bandit problem~\cite{langford2007epoch,li2010contextual}. 
In contextual bandit problem, at each iteration a learner is faced with $m$ arms associante with feature vectors $x_1, \dots, x_m$, and the learner is trying to pull an arm at each round to optimize the overall reward. 
In our case, $m$ arms correspond to the $m$ sentences $s_{(1)}, \dots, s_{(m)}$ in a prompt, and features can be obtained by $\phi(s_{(i)})$ where $\phi$ is a text encoder.  Utilizing feature information is crucial because if modifying a particular sentence can lead to performance improvement, it is likely that modifying similar sentences will also yield positive results. 

We then adopt the Lin-UCB algorithm~\cite{li2010contextual}, a widely recognized algorithm for contextual bandit problems, to guide sentence selection. Given the history $\{(s_t^\text{before}, s_t^\text{after}, r_t)\}_{t=1}^{T-1}$, we compute a linear estimator of the underlying reward model
\begin{equation}
    w_T^* = (H^T H + \lambda I)^{-1}H^T r, 
\end{equation}
where $H$ is a $(T-1)$-by-$d$ matrix with each row being $\phi(s_t^\text{before})$, and $r = [r_1, \dots, r_{T-1}]^T$. 
This is simply a solution of a ridge regression with feature matrix $H$ and reward $r$.
The following UCB rule is then used to select the sentence for mutation: 
\begin{equation}
    \arg\max_i   \phi(s_{(i)})^T w_T^* + \alpha \sqrt{\phi(s_{(i)})^T A^{-1}\phi(s_{(i)})}, 
    \label{eq:linucb}
\end{equation}
where $A = H^T H + \lambda I$. Since this Lin-UCB estimation may not be accurate, we let the algorithm has probability $P$ to choose a arm by purely random and $1-P$ to choose the sentence based on \eqref{eq:linucb}.

\section{Experimental Results}
\label{sec:exp}

In this section, we present empirical evidence demonstrating that the proposed long prompt tuning method can significantly enhance performance on the Big-Bench Hard (BBH) benchmark~\cite{suzgun2022challenging,srivastava2022beyond}. We then conduct ablation studies to analyze the proposed algorithm's effectiveness and show qualitative results on how the proposed algorithm refines human-written prompts.

\subsection{Experimental Settings}

We consider the prompt developed in \cite{suzgun2022challenging} for the BBH tasks, where prompts consist of two parts: Task Description and Demos. The Task Description provides instructions describing the task, while each demo includes the question, a chain-of-thoughts demonstration illustrating the problem-solving process step-by-step, and the final answer. During prompt tuning, the format (e.g., ``Question. '', ``Answer. '')  are retained, while the tuning algorithm is allowed to modify any other parts including the instruction part and the chain-of-thoughts part. Table \ref{tab:statistics} summarizes the statistics of the datasets used in this study. The full prompts for each data can be downloaded from \url{https://github.com/suzgunmirac/BIG-Bench-Hard}.

Each task consists of 250 samples, randomly divided into 50\% for training and 50\% for testing. Only training samples are utilized for prompt tuning. All tasks are multi-class classification problems, and we report accuracy on training data, test data, and the combined dataset when comparing different tuning methods.

 For the main experiments, we evaluated the performance on the text-bison model and utilized the instruction-tuned PaLM 2-L model as the LLM mutator. Both models belong to the PaLM 2-model family~\cite{anil2023palm}. Following the previous prompt tuning works, we set temperature as $0$. However, to enhance the diversity of sentence mutation, we set the temperature of the LLM-Mutator to $0.5$. 

\begin{table}[]
     \centering
     \resizebox{0.5\textwidth}{!}{%
  
\begin{tabular}{|c|c|c|}
\hline
\multirow{2}{*}{Dataset} & \multirow{2}{*}{Number of Words} & Number of \\ 
& & Mutable Sentences \\\hline
Causal Judgement & 678 & 20 \\ \hline
Salient Translation & 745 & 34 \\ \hline
Disambiguation & 671 & 37 \\ \hline
Formal Fallacies & 741 & 32 \\ \hline
Hyperbaton & 539 & 19 \\ \hline
Dyck Language & 679 & 22 \\ \hline
Color Reasoning & 435 & 17 \\ \hline
Logical Five & 461 & 25 \\ \hline
\end{tabular}
     }
     \caption{Initial prompt statistics for each dataset. When calculating the ``number of mutable sentences,'' we only consider sentences that can be modified during the search phase. }
     \label{tab:statistics}
 \end{table}

 \begin{table*}[]
     \centering
     \resizebox{1\textwidth}{!}{%
\begin{tabular}{|c|c|c|c|c|c|c|c|c|c|}
\hline
\multirow{2}{*}{Task} & {Original} & \multicolumn{2}{c|}{Genetic Algorithm} & \multicolumn{2}{c|}{Evolve ``Step-by-step''} &\multicolumn{2}{c|}{Greedy} & \multicolumn{2}{c|}{Our Method}\\ \cline{3-10}
 & Prompt  & Train & Test & Train & Test & Train & Test & Train & Test \\ \hline
Causal Judgment & 58.9  & 63.1$\pm$1.3 & 59.2$\pm$0.5 & 63.1$\pm$0.5 & 58.2$\pm$0.5 & 63.0$\pm$1.0 & 59.6$\pm$2.9  &  67.7$\pm$1.5&  63.7$\pm$1.3  \\ \hline
Salient Translation & 54.4  & 58.7$\pm$2.0  & 56.5$\pm$4.3  & 57.1$\pm$0.4  & 55.7$\pm$0.4 & 59.5$\pm$1.5 & 53.6$\pm$3.3 &  60.0$\pm$3.5 & 58.7$\pm$2.6 \\ \hline
Disambiguation & 58.0  & 64.8$\pm$1.6  & 61.0$\pm$0.7  &  64.6$\pm$1.5 & 61.8$\pm$1.0 & 66.7$\pm$0.6 &63.4$\pm$2.7  & 70.1$\pm$1.9  & 68.0$\pm$2.1 \\ \hline
Formal Fallacies & 60.0  & 63.5$\pm$1.1  & 64.5$\pm$0.7 &60.6$\pm$1.0 & 62.4$\pm$1.7  & 68.5$\pm$2.6  & 68.8$\pm$3.0 & 70.4$\pm$3.9 & 73.1$\pm$3.6 \\ \hline
Hyperbaton & 74.4 & 82.8$\pm$1.3 & 76.3$\pm$2.5  & 84.9$\pm$0.0 & 79.6$\pm$0.3 & 83.5$\pm$3.3 & 79.3$\pm$5.3  &  88.4$\pm$2.6 & 85.5$\pm$1.1 \\
\hline
Dyck Language &16.0  & 20.2$\pm$1.7  & 16.4$\pm$0.4   & 18.25$\pm$0.0  & 20.4$\pm$0.1  &19.0$\pm$0.4   & 17.3$\pm$0.6 & 22.8$\pm$0.7 & 20.9$\pm$0.4 \\ \hline
Color Reasoning  & 82.0  & 85.1$\pm$0.4  & 78.9$\pm$1.7  & 86.2$\pm$0.7 & 81.5$\pm$1.3 & 85.1$\pm$0.7 & 81.1$\pm$0.4  & 85.4$\pm$0.4  & 83.5$\pm$0.4 \\ \hline
 Logical Five & 38.8 &  49.3$\pm$2.7 & 48.2$\pm$1.6  & 54.5$\pm$1.0  & 50.3$\pm$0.4  & 53.9$\pm$3.9 & 51.3$\pm$1.2 &  59.8$\pm$2.2& 54.7$\pm$0.9 \\ \hhline{==========} 
 Average & 55.3 & 60.94 & 57.6  & 61.2 & 58.7 & 62.4 & 59.3 & 65.6 &  63.5 \\ \hline
\end{tabular}

     }
     \caption{Main results on BBH benchmark. The search budget is limited to 50 evaluations over the training set for each method. We ran each experiment 3 times and report the mean and standard deviation.  }
     \label{tab:main_results}
 \end{table*}

\subsection{Main results}
 We compare the proposed algorithm with the following baselines: 
\begin{itemize}
    \item Original Prompt: Performance of the original human-designed prompt developed in~\cite{suzgun2022challenging}, which also serves as the initialization for other tuning methods.
    \item Genetic Algorithm: An implementation of the genetic algorithm for long prompt tuning. To facilitate a more comparable comparison with our method, we set the pool size to $4$. At each step, $8$ new candidates are generated by randomly mutating and performing crossover on the top candidates within the pool.
    \item Evolve ``step-by-step'': Several recent prompt tuning studies have explored the concept of evolving a prompt using a single sentence, such as the 'Let's think step by step' sentence employed in chain-of-thought prompts~\cite{kojima2022large}. We adopted one of the state-of-the-art methods for single sentence optimization~\cite{yang2023large} to optimize all of the 'Let's think step by step' sentences within the chain-of-thought prompt.
    \item Greedy: A simple greedy approach mentioned in Section~\ref{sec:framework}, where we only store the top performing candidate in the pool and generating new candidate on top of it. 
\end{itemize}
For our method, we use the same hyper-parameters for all the experiments.
We set the pool size (beam size) to $4$ for all experiments. When applying guided mutation in Section~\ref{sec:guided}, we use a T5 encoder to encode both the current entry and the history and normalize the embeddings to have a unit $\ell_2$ norm. We retrieve only the top $4$ history entries and require the $\ell_2$ distance between the encoded history and the current sentence to be below $0.5$. When applying the sentence selection algorithm described in Section~\ref{sec:guided}, we set $\alpha=0.05$ in~\eqref{eq:linucb} and set $P=0.5$. Note that those hyper-parameters are not well tuned.

%

For this experiment, we limited the computational budget to 50 evaluations on the training set and reported the training, test, and combined accuracy achieved by each method. All the experiments are run three times and we report the mean and standard deviation. The results are summarized in Table~\ref{tab:main_results}. Our proposed algorithm outperformed the other methods, demonstrating significant improvements in accuracy across all tasks. We also compute the improvements of accuracy on the whole dataset (including both training and testing) and report the accuracy gain in Figure~\ref{fig:example_1}. 

Across all 8 tasks, our algorithm achieves an average of 8.2\% gain in test accuracy and 9.2\% gain in the accuracy of full evaluation set (train + test). Among these tasks, we achieve the largest performance gain (18.45\%) on the logical deduction task, and the smallest gain (2.45\%) on reasoning about colored objects (Color Reasoning). One potential reason of small performance gain on color reasoning is that the original prompt already achieves very high accuracy, so there is not much room for improvements. 

Comparing the baselines, it becomes evident that evolving a single sentence (Evolve 'step-by-step') fails to achieve substantial improvements in long prompt tuning. This is mainly due to the fact that long prompt tuning typically involves over 20 sentences with detailed instructions and explanations, and thus naively modifying ``Let's think step by step'' is unlikely to lead to significant enhancements. Furthermore, the Genetic Algorithm and Greedy Algorithm each exhibit their own limitations, as discussed in Section~\ref{sec:framework}, which are reflected in their inferior performance compared to our method.

Despite being able to significantly boost the performance, we also observe some degree of overfitting in our search procedure. Although the proposed beam search method can partially mitigate overfitting (see Section~\ref{sec:framework}), we still observe higher training accuracy than testing in most cases. However, as the performance gain is substantial, the prompts found by the algorithm are still significantly better than the original one, even on the test set.

\subsection{Ablation Study}
We conduct an ablation study on the two techniques  introduced in Section~\ref{sec:guided}: the history-guided mutation and the contextual bandit algorithm for sentence selection. The results are presented in Table~\ref{tab:ablation}. We can observe that both components are contributing to the final performance of the model.

\begin{table*}[]
     \centering
     \resizebox{0.9\textwidth}{!}{%
     \begin{tabular}{|c|c|c|c|c|c|c|c|c|}
     \hline
        \multirow{2}{*}{Task} &  \multicolumn{2}{c|}{Disambiguation} & \multicolumn{2}{c|}{Formal Fallacies} & \multicolumn{2}{c|}{Hyperbaton} &\multicolumn{2}{c|}{Logical Five} \\ \cline{2-9}
        & Train & Test& Train & Test& Train & Test& Train & Test\\
        \hline
     Our method (no history-guided mutation)  &  67.4  & 66.9 & 61.1  & 66.1 &  83.6 & 78.1 & 55.6 & 53.2  \\
     \hline
     Our method (no sentence selection) &  68.2 & 63.1 & 71.2 & 72.8 & 81.4 & 80.3 & 54.2 & 50.4  \\
     \hline
     Our method &  70.1 & 68.0  & 70.4 & 73.1 & 88.4 &  85.5 & 59.8 & 54.7  \\
     \hline
    
     \end{tabular}     }
     \caption{Ablation Study on the two components introduced in Section~\ref{sec:guided}. }
     \label{tab:ablation}
 \end{table*}

\begin{table*}[ht]
\centering
\adjustbox{max width=1\textwidth}{
\begin{tabular}{|c|c|}
\hline
\bf 
Original prompt: 38.8\% accuracy &
\bf New prompt found at iteration 48: 56\% accuracy (train 57.9\% / test 54.0\%) \\
\hline
A: Let's think step by step. &
A: Let's think things through one step at a time. \\
\hdashline
(2) Eli finished below Amy: "(above) ? Amy ? Eli ? (below)". &
(2) Amy was above Eli: "(above) ? Amy ? Eli ? (below)".\\
\hdashline
 Eli finished last. &
 Eli came in last place. \\
 \hdashline
 \makecell{Q: The following paragraphs each describe a\\ set of three objects arranged in a fixed order.\\ The statements are logically consistent within each\\ paragraph. On a shelf, there are three books: a white\\ book, a green book, and an orange book. The green \\ book is to the right of the white book. The orange book is the rightmost.} &
\makecell{Q: Each paragraph below describes three objects \\ arranged in a fixed order, and the statements are logically consistent in each paragraph.\\ There are three books on a shelf: a white, green, and orange book.\\ The green book is to the right of the white book, and \\the orange book is in the far right position.} \\
 \hdashline
 \makecell{A: Let's think step by step.} &
\makecell{
A: Let's think through things one step at a time.
} \\
 \hdashline
 \makecell{The white book is the leftmost.} &
\makecell{The white book is at the far left.
} \\
 \hdashline
\makecell{A: Let's think step by step.} &
\makecell{A: Let's think one step at a time.
 }\\
\bottomrule
\end{tabular}}
\caption{An example that our method improves the performance on Logical Deduction Five Objects from 38.8\% to 56\%. 
}
\label{tab:example_logic}
\end{table*}

\begin{table*}[ht]
\centering
\adjustbox{max width=1\textwidth}{
\begin{tabular}{|c|c|}
\hline
\bf 
Original promp: 60\% accuracy &
\bf New prompt found at iteration 91: 76\% accuracy  (train 92.1\%, test 83.1\%) \\
\hline
Distinguish deductively valid arguments from formal fallacies. &
{ Identify deductively valid arguments from formal fallacies.} \\
\hdashline
So, it is true that Lesley is a great-grandfather of Leroy. &
{ So it's true that Lesley is Leroy's great-grandfather.  } \\
\hdashline
 \makecell{Whoever is not a great-grandfather of Clyde is a stepbrother \\ of Brian: If X = NOT (great-grandfather(Clyde)), then \\ X = stepbrother(Brian).} &
 \makecell{* If someone is a stepbrother of Brian then they \\ are not a great-grandfather of Clyde.} \\
 \hdashline
 \makecell{Furthermore, by (1), we have if X = NOT \\ (great-grandfather(Clyde)), then X = stepbrother(Brian).} &
\makecell{Additionally, according to (1) we have that if X = NOT \\ (great-grandfather(Clyde)), then X = stepbrother(Brian).} \\
 \hdashline
 \makecell{By the transitive relation rule in first-order logic, \\we then have: if X = ancestor(Dana), then X = stepbrother(Brian).} &
\makecell{
Using the transitive relation rule in first-order logic, we have: \\if X = ancestor(Dana), then X = stepbrother(Brian).
} \\
 \hdashline
 \makecell{Let’s see whether the Hypothesis can be deduced from \\ the arguments (1) and (2) by logical reasoning?} &
\makecell{Let’s see whether the Hypothesis is a logical  consequence \\of the arguments (1) and (2)? } \\
 \hdashline
\makecell{So, from (1) and (2), we cannot necessarily deduce \\ the Hypothesis.} &
\makecell{So, given (1) and (2), we are not always entitled to infer \\ the Hypothesis.  }\\
\bottomrule
\end{tabular}}
\caption{An example demonstrating how our method changes the original prompt. By conducting these changes the combined accuracy on Formal Fallacies can be improved from 60\% to 76\%. Interestingly, there's one line
}
\label{tab:example_fallacies}
\end{table*}

\subsection{Qualitative results}
\label{sec:qualitative}
One important benefit of automatic hard prompt engineering is that the resulting prompts remain interpretable by humans, allowing users to easily verify the modifications. We provide some successful examples found by our search in Table~\ref{tab:example_logic} and Table~\ref{tab:example_fallacies}. In each table, we  show only the different parts of the initial prompt from BBH and the changes made by our method. 

The first example demonstrated in Table~\ref{tab:example_logic} is for the logic deduction task on five objects. The initial prompt achieves 38.8\% accuracy while the revised prompt found at iteration 48 improves the performance to 57.9\% train accuracy and 54.0\% test accuracy. We observed that most of the changes involve minor revisions to the original sentence without altering its meaning. These seemingly insignificant modifications can lead to substantial improvements in LLM accuracy, showcasing the important of automatic long prompt engineering.  

The second prompt in Table~\ref{tab:example_fallacies} is for the Formal Fallacies task, designed to identify whether a given logic statement is valid. The initial prompt achieves 60\% accuracy while the revised prompt found at iteration  91 improves the train accuracy to 92.1\% and test accuracy to 83.1\%. Similar to the previous case, most of the changes  involve minor revisions. In this case, we also want to highlight  a potential limitation of the proposed method that could be addressed in future work. In the sentence marked as *, the revised sentence is {\it not} semantically equivalent to the original one. The original and revised sentences represent different logical statements (original sentence: not A $\rightarrow$ B; new sentence: B $\rightarrow$ not A). However, the LLM appears incapable of detecting this subtle distinction, leading to an erroneous rephrasing. Although this change leads to improve training accuracy, it actually hurts test accuracy. Therefore, incorrect mutations can lead to overfitting, and developing strategies to mitigate these errors during the search process would be an interesting area of future research.

\section{Conclusions, Limitations, and Future Work}
We study the problem of automatic prompt engineering for long prompts, often comprising thousands of tokens. We investigate the performance of the standard greedy algorithm and genetic algorithm, and develop  a novel search algorithm that yields superior performance. With only 50 evaluations on the training set, our method achieves an average absolute accuracy improvement of 9.2\% across 8 tasks from Big Bench Hard. This demonstrates the significant potential benefits of automatic long prompt tuning and underscores the importance of this emerging area.

As the first paper focusing on automatic engineering of  the entire long prompts, we identify several limitations of the current methods which can lead to interesting future research: 
\begin{itemize}
    \item  The current algorithm relies on using  another LLM to rephrase a sentence. As illustrated  in Section~\ref{sec:qualitative}, this LLM-mutator may introduce errors during sentence rewriting, particularly for intricate sentences (e.g., CoT in complicated logical deduction tasks). Therefore, improving the ``correctness'' of LLM-Mutator is an interesting future area of research, which has not been fully addressed in our work as well as other recent studies~\cite{fernando2023promptbreeder,guo2023connecting}. 
    \item Similar to any other training or tuning algorithms, automatic prompt engineering can suffer from overfitting to the training data, as discussed in Section~\ref{sec:exp}.  This overfitting can be potentially alleviated by applying additional regularization techniques, such as imposing sparsity constraints by reducing the number of modified sentences. However, further research is required to identify the most effective regularization strategies for automatic prompt engineering.
 
    \item In the current implementation, we break down the long prompt into individual sentences and modify one sentence at a time. However, it might be beneficial to manipulate multiple sentences simultaneously during mutation or consolidate multiple sentences into a single one. An automated mechanism for carrying out this process would be an interesting direction for enhancing our method.   
    
    \item Although our algorithm is able to find a good solution with less than 100 evaluations on the  training set, the cost is still not negligible especially when tuning prompts using APIs with cost or rate limits. Employing early stopping techniques, where the evaluation of poorly performing candidates are terminated early, could potentially reduce the number of queries.
    
    \item Although automatic prompt engineering can achieve significant gain,
    the search space of hard prompt has limited representation power which hinders further performance improvements. 
    It has been shown that soft prompt tuning has limited representation power~\cite{wang2023universality}, and our search space is a small subset of soft prompts. 
     Therefore, when provided with sufficient data, computational resources, and white-box access to the LLM, (parameter-efficient) fine-tuning may still achieve superior performance.
\end{itemize}

\section*{Acknowledgements}
We thank Ruochen Wang, Vineet Gupta, Kedar Dhamdhere, Daliang Li for valuable discussions and feedback.

\newpage 

\bibliography{tacl} 
\bibliographystyle{acl_natbib}
  
\end{document}

\begin{table*}
    \begin{tabular}{|c|c|c|c|c|c|c|c|}
     \hline
          & Causal Judgement & Salient Translation & Disambiguation & formal fallacies & Hyperbaton & Boolean Expression\\
          \hline
     Original Prompt  & 0.581 / 0.596   & 0.544 / 0.544 & 0.587 / 0.573 &  0.579 / 0.621 & 0.770 / 0.718 & 0.881 / 0.879  \\
     \hline
     Genetic Algorithm & 0.634 / 0.596  & 0.568 / 0.576 & 0.635 / 0.653 & 0.69 / 0.80 (?) &  0.842 / 0.750 & 0.905 / 0.903 \\
     \hline
     Evolve ``Step-by-step''  & 0.613 / 0.618   & 0.544 / 0.544 & 0.587 / 0.573 & 0.587 / 0.596  & 0.841 / 0.766 & 0.905 / 0.887 \\
     \hline
       Greedy & 0.624 / 0.596  & 0.584 / 0.544  & 0.643 / 0.605 & 0.706 / 0.685 & 0.849 / 0.815 & 0.905 / 0.887 \\
     \hline
     Our method & 0.656 / 0.638  & 0.616 / 0.592 &0.698 / 0.734  & 0.706 / 0.693 & 0.897 / 0.871 & 0.905 / 0.895  \\
     \hline
    
     \end{tabular}
\end{table*}

\begin{table*}[ht]
\centering
\begin{tabular}{@{}m{0.52\textwidth}@{}}
\toprule
 \setrow{\bfseries} Original prompt (0.579 / 0.621 ) \\
 Changes by our method (0.802 / 0.798) \\
\hline
Distinguish deductively valid arguments from formal fallacies. \\
{\bf Identify deductively valid arguments from formal fallacies.} \\
So, it is true that Lesley is a great-grandfather of Leroy.
{\bf So it's true that Lesley is Leroy's great-grandfather.  } \\
 Whoever is not a great-grandfather of Clyde is a stepbrother of Brian: If X = NOT (great-grandfather(Clyde)), then X = stepbrother(Brian).\\
{\bf * If someone is a stepbrother of Brian then they are not a great-grandfather of Clyde.} \\
 Furthermore, by (1), we have if X = NOT (great-grandfather(Clyde)), then X = stepbrother(Brian).\\
{\bf Additionally, according to (1) we have that if X = NOT (great-grandfather(Clyde)), then X = stepbrother(Brian).} \\
 By the transitive relation rule in first-order logic, we then have: if X = ancestor(Dana), then X = stepbrother(Brian). \\
{\bf 
Using the transitive relation rule in first-order logic, we have: if X = ancestor(Dana), then X = stepbrother(Brian).
} \\
 Let’s see whether the Hypothesis can be deduced from the arguments (1) and (2) by logical reasoning? \\
{\bf Let’s see whether the Hypothesis is a logical  consequence of the arguments (1) and (2)? } \\
 \\
 So, from (1) and (2), we cannot necessarily deduce the Hypothesis.
{\bf So, given (1) and (2), we are not always entitled to infer the Hypothesis.  }\\
\bottomrule
\end{tabular}
\caption{An example demonstrating how our method changes the original prompt. By conducting these changes the combined accuracy on Formal Fallacies can be improved from 60\% to 80\% (20\% improvement). Interestingly, there's one line
}
\label{tab:example_fallacities}
\end{table*}

 \begin{table*}[]
     \centering
     \resizebox{1\textwidth}{!}{%
\begin{tabular}{|c|c|c|c|c|c|c|c|c|c|c|}
\hline
\multirow{2}{*}{Task} & \multicolumn{2}{c|}{Original Prompt} & \multicolumn{2}{c|}{Genetic Algorithm} & \multicolumn{2}{c|}{Evolve ``Step-by-step''} &\multicolumn{2}{c|}{Greedy} & \multicolumn{2}{c|}{Our Method}\\ \cline{2-11}
 & Train & Test & Train & Test & Train & Test & Train & Test & Train & Test\\ \hline
Causal Judgment & 58.1 & 59.6  & 63.1$\pm$1.3 & 59.2$\pm$0.5 & 63.1$\pm$0.5 & 58.2$\pm$0.5 & 63.0$\pm$1.0 & 59.6$\pm$2.9  &  67.7$\pm$1.5&  63.7$\pm$1.3 \\ \hline
Salient Translation & 54.4 & 54.4  & 58.7$\pm$2.0  & 56.5$\pm$4.3  & 57.1$\pm$0.4  & 55.7$\pm$0.4 & 59.5$\pm$1.5 & 53.6$\pm$3.3 &  60.0$\pm$3.5 & 58.7$\pm$2.6 \\ \hline
Disambiguation & 58.7 &57.3  & 64.8$\pm$1.6  & 61.0$\pm$0.7  &  64.6$\pm$1.5 & 61.8$\pm$1.0 & 66.7$\pm$0.6 &63.4$\pm$2.7  &  & \\ \hline
Formal Fallacies & 57.9 & 62.1  & 63.5$\pm$1.1  & 64.5$\pm$0.7 &60.6$\pm$1.0 & 62.4$\pm$1.7  & 68.5$\pm$2.6  & 68.8$\pm$3.0 & 70.4$\pm$3.9 & 73.1$\pm$3.6 \\ \hline
Hyperbaton & 77.0 & 71.8  & 82.8$\pm$1.3 & 76.3$\pm$2.5  & 84.9$\pm$0.0 & 79.6$\pm$0.3 & 83.5$\pm$3.3 & 79.3$\pm$5.3  &  88.4$\pm$2.6 & 85.5$\pm$1.1 \\
\hline
Dyck Language & 13.5 & 18.5  & 20.2$\pm$1.7  & 16.4$\pm$0.4   & 18.3$\pm$0.0  & 20.4$\pm$0.1  &19.0$\pm$0.4   & 17.3$\pm$0.6 & 22.8$\pm$0.7 & 20.9$\pm$0.4 \\ \hline
Color Reasoning  & 82.5& 81.5  & 85.1$\pm$0.4  & 78.9$\pm$1.7  & 86.2$\pm$0.7 & 81.5$\pm$1.3 & 85.1$\pm$0.7 & 81.1$\pm$0.4  & 85.4$\pm$0.4  & 83.5$\pm$0.4 \\ \hline
 Logical Five & 37.3& 40.3 &  49.3$\pm$2.7 & 48.2$\pm$1.6  & 54.5$\pm$1.0  & 50.3$\pm$0.4  & 53.9$\pm$3.9 & 51.3$\pm$1.2 &  59.8$\pm$2.2& 54.7$\pm$0.9 \\ \hline
\end{tabular}

     }
     \caption{Main results on BBH benchmark. The search budget is limited to 50 evaluations for each method. }
     \label{tab:main_results}
 \end{table*}

 \begin{table*}[]
     \centering
     \resizebox{1\textwidth}{!}{%
\begin{tabular}{|c|c|c|c|c|c|c|c|c|c|c|}
\hline
\multirow{2}{*}{Task} & \multicolumn{2}{c|}{Original Prompt} & \multicolumn{2}{c|}{Genetic Algorithm} & \multicolumn{2}{c|}{Evolve ``Step-by-step''} &\multicolumn{2}{c|}{Greedy} & \multicolumn{2}{c|}{Our Method}\\ \cline{2-11}
 & Training & Test & Training & Test & Training & Test & Training & Test & Training & Test\\ \hline
Causal Judgment & 0.581 & 0.596 & 0.634 & 0.596 & 0.613 & 0.618 & 0.624 & 0.596 & 0.656 & 0.638 \\ \hline
Salient Translation & 0.544 & 0.544 & 0.568 & 0.576 & 0.544 & 0.544 & 0.584 & 0.544 & 0.616 & 0.592  \\ \hline
Disambiguation & 0.587 & 0.573 & 0.635 & 0.653 & 0.587 & 0.573 & 0.643 & 0.605 & 0.698 & 0.734 \\ \hline
Formal Fallacies & 0.579 & 0.621 & 0.69 & 0.80 (?) & 0.587 & 0.596 & 0.706 & 0.685 & 0.675 & 0.766 \\ \hline
Hyperbaton & 0.770 & 0.718 & 0.842 & 0.750 & 0.841 & 0.766 & 0.849 & 0.815 & 0.897 & 0.871 \\ \hline
Boolean Expression & 0.881 & 0.879 & 0.905 & 0.895 & 0.905 & 0.887 & 0.905 & 0.887 & 0.905 & 0.895 \\ \hline

\end{tabular}

     }
     \caption{Main results on BBH benchmark. The search budget is limited to 50 evaluations for each method. }
     \label{tab:main_results}
 \end{table*}